\documentclass[twoside]{article}
\usepackage{nips15submit_e,times}
\usepackage{hyperref}
\usepackage{url}
\usepackage{graphicx}
\usepackage{wrapfig}
\usepackage{subcaption}
\usepackage{mwhmacros}
\usepackage{array}
\usepackage{todonotes}

\def\D{\calD}

\def\x{\vx}

\def\xrec{\hat\vx}
\def\xstar{\vx_\star}
\def\fstar{f_\star}

\newcounter{xxx}
\begin{document}

\title{Unbounded Bayesian Optimization via Regularization}
\author{
Bobak Shahriari\\
University of British Columbia\\
\texttt{bshahr@cs.ubc.ca}\\
\And
Alexandre Bouchard-C\^ot\'e\\
University of British Columbia\\
\texttt{bouchard@stat.ubc.ca}\\
\And
Nando de Freitas\\
University of Oxford\\
Google DeepMind\\
\texttt{nando@google.com}\\
}

\nipsfinalcopy

\maketitle

\begin{abstract}
Bayesian optimization has recently emerged as a popular and efficient tool for
global optimization and hyperparameter tuning.
Currently, the established Bayesian optimization practice requires a
user-defined bounding box which is assumed to contain the optimizer.
However, when little is known about the probed objective function, it can be
difficult to prescribe such bounds.
In this work we modify the standard Bayesian optimization framework in a
principled way to allow automatic resizing of the search space. 
We introduce two alternative methods and compare them on two common synthetic
benchmarking test functions as well as the tasks of tuning the stochastic
gradient descent optimizer of a multi-layered
perceptron and a convolutional neural network on MNIST.
\end{abstract}

%===============================================================================
\section{Introduction}
\label{sec:introduction}

Bayesian optimization has recently emerged as a powerful tool for the
efficient global optimization of black box objective functions.
Since the technique was introduced over 50 years ago, it has been applied for
many different applications. Perhaps the most relevant application for the
machine learning community is the automated tuning of algorithms~\cite{Bergstra:2011,Mahendran:2012,Snoek:2012,Swersky:2013,Hoffman:2014}.
However, the current state of the art requires the user to prescribe a bounding
box within which to search for the optimum.
Unfortunately, setting these bounds---often arbitrarily---is one of the main
difficulties hindering the broader use of Bayesian optimization as a standard
framework for hyperparameter tuning.
For example, this obstacle was raised at the NIPS 2014 Workshop on Bayesian
optimization as one of the open challenges in the field. 

In the present work, we propose two methods that are capable of growing the 
search space as the optimization progresses.
The first is a simple heuristic which regularly doubles the volume of the
search space, while the second is a regularization method that is practical
and easy to implement in any existing Bayesian optimization toolbox based on
Gaussian Process priors over objective functions.

At a high level, the regularized method is born out of the observation that
the only component of Bayesian optimization that currently requires a
bounding box on the search space is the maximization of the acquisition
function; this constraint is necessary because acquisition functions can
have suprema at infinity. By using a non-stationary prior mean
as a regularizer, we can exclude this possibility and use an unconstrained
optimizer, removing the need for a bounding box.
%In a second step, we provide an equivalent view that involves a structured prior mean on the Gaussian Process. With this
%view, we are able to experiment with methods that are not improvement based,
%such as entropy search~\cite{Villemonteix:2009,Hennig:2012,Hernandez:2014} and
%Thompson sampling~\cite{Hernandez:2014}.

\subsection{Related work}

Although the notion of using a non-trivial Gaussian process prior mean is not
new, it is usually expected to encode domain expert knowledge or known
structure in the response surface.
Only one recent work, to the best of the authors' knowledge, has considered
using the prior mean as a regularization term and it was primarily to avoid
selecting points along boundaries and in corners of the bounding
box~\cite{Snoek:2015DNGO}.

In this work we demonstrate that regulization can be used to carry out
Bayesian optimization without a rigid bounding box.
We further introduce a \emph{volume doubling} baseline, which we compare
to the regularization approach.
While the regularized algorithms exhibit a much more homogeneous search
behaviour (\emph{i.e.}\ boundaries and corners are not disproportionately
favoured), the volume doubling baseline performs very well in practice.

We begin with a brief review of Bayesian optimization with Gaussian processes
in the next section, followed by an introduction to regularization via
non-stationary prior means in Section~\ref{sec:unbounded}, including
visualizations that show that our proposed approaches indeed venture out of the
initial user-defined bounding box.
Section~\ref{sec:experiments} reports our results on two synthetic benchmarking
problems and two neural network tuning experiments.

%===============================================================================
\section{Bayesian optimization}
\label{sec:method}

Consider the global optimization problem of finding a global maximizer
$\xstar\in\argmax_{\x\in\R^d} f(\x)$ of an unknown objective function
$f:\R^d\mapsto\R$.
In this work, the function $f$ is assumed to be a black-box for which we have
no closed-form expression or gradient information.
We further assume that $f$ is expensive to evaluate so we wish to locate the
best possible input $\vx$ with a relatively small budget of $N$ evaluations.
Finally, the evaluations $y\in\R$ of the objective function are noise-corrupted
observations, and
for the remainder of this work, we assume a Gaussian noise distribution,
$y\sim \Normal(f(\x), \sigma^2)$.
In contrast to typical Bayesian optimization settings, we do not assume the $\argmax$
to be restricted to  a bounded subset $\calX\subset\R^d$. 

Bayesian optimization is a sequential model-based approach which involves (i)
maintaining a probabilistic \emph{surrogate} model over likely functions given
observed data; and (ii) selecting future query points according to a selection
\emph{policy}, which leverages the uncertainty in the surrogate to negotiate
exploration of the search space and exploitation of current modes.
The selection policy is represented by an \emph{acquisition function}
$\alpha_n:\R^d\mapsto\R$, where the subscript indicates the implicit dependence
on the surrogate and, by extension, on the observed data
$\D_n=\{(\vx_i,y_i)\}_{i=1}^n$.
More precisely, at iteration $n$: an input $\vx_{n+1}$ is selected by
maximizing the acquisition function $\alpha_n$; the black-box is queried
and produces a noisy $y_{n+1}$; and the surrogate is updated in light of the 
new data point $(\vx_{n+1}, y_{n+1})$.
Finally, after $N$ queries the algorithm must make a final recommendation
$\xrec_N$ which represents its best estimate of the optimizer.

\subsection{Gaussian processes}

In this work we prescribe a Gaussian process (GP) prior over
functions~\cite{Rasmussen:2006}.
When combined with a Gaussian likelihood, the posterior is also a GP and the
Bayesian update can be computed analytically.
Note that other surrogate models, such as random forests, have also been
proposed in the model-based optimization literature~\cite{Hutter:2010b}.

A Gaussian process $\mathrm{GP}(\mu_0, k)$ is fully characterized by its prior
mean function $\mu_0:\R^d\mapsto\R$ and its positive-definite kernel, or
covariance, function $k:\R^d\times\R^d\mapsto\R$. Given any finite
collection\footnote{Here we use the convention $a_{i:j} = \{a_i,\dots,a_j\}$.}
of $n$ points $\x_{1:n}$, the values of $f(\vx_1),\dots,f(\vx_n)$ are jointly
Gaussian with mean $\vm$, where $m_i:=\mu_0(\vx_i)$, and $n\times n$
covariance matrix $\vK$, where $K_{ij}=k(\x_i,\x_j)$---hence the term
covariance function.

Given the Gaussian likelihood model, the vector of concatenated observations
$\vy=y_{1:n}$ is also jointly Gaussian with covariance $\vK+\sigma^2\vI$.
Therefore, at any arbitrary test location $\x$, we can query our surrogate
model (the GP) for the function value $f(\x)$ conditioned on observed data
$\D_n$.
The quantity $f(\x)$ is a Gaussian random variable with the following mean
$\mu_n(\x)$ and marginal variance $\sigma^2_n(\x)$
\begin{align}
    \mu_n(\x)
    &= \mu_0(\x) + \vk(\x)\T(\vK+\sigma^2\vI)^{-1}(\vy - \vm)\,,
    \label{eq:gpmean}
    \\
    \sigma^2_n(\x)
    &= k(\x,\x) - \vk(\x)\T(\vK+\sigma^2\vI)^{-1}\vk(\x)\,,
    \label{eq:gpvar}
\end{align}
where $\vk(\x)$ is the vector of cross-covariance terms between $\x$ and
$\x_{1:n}$.
Throughout this work we use the squared exponential kernel
\begin{align}
    k_\textsc{se}(\vx, \vx')
	&= \theta_0 \exp(-\tfrac12r^2),
    % \\
%    k_\textsc{mat\'ern}(\vx,\vx')
%    &= \theta_0 \exp(-\sqrt5 r) (1+\sqrt{5}r+\tfrac53r^2).
\end{align}
where $r = (\vx-\vx')\T\vLambda^{-1}(\vx-\vx')$ and $\vLambda$ is a diagonal
matrix of $d$ length scales $\theta_{1:d}$ and $\theta_0$ is the kernel
amplitude.
Collectively referred to as the \emph{hyperparemeter}, $\vtheta = \theta_{0:d}$
parameterizes the kernel function $k$. When the noise variance $\sigma^2$ is
unknown, it can be added as a model hyperparameter as well. Similarly, the most
common agnostic choice of prior mean is a constant bias $\mu_0(\vx)\equiv b$
which can also be added to $\vtheta$. 

\paragraph{Hyperparameter marginalization.}
As in many regression tasks, the hyperparameter $\vtheta$ must somehow be specified
and has a dramatic effect on performance.
Common tuning techniques such as cross-validation and maximum likelihood are
either highly data-inefficient or run the risk of overfitting.
Recently, a Bayesian treatment of the hyperparameters via Markov chain Monte Carlo
has become standard practice in Bayesian optimization~\cite{Snoek:2012}.
Similarly in the present work, we specify an uninformative prior on $\vtheta$ and
approximately integrate it by sampling from the posterior $p(\vtheta|\D_n)$ via
slice sampling.
%Alternatively one can use a sequential Monte Carlo technique as
%in~\cite{Gramacy:2011}.

\subsection{Acquisition functions}

So far we have described the statistical model we use to represent our belief
about the unknown objective $f$ given data $\D_n$, and how to update this belief
given new observations.
We have not described any mechanism or policy for selecting the query
points $\x_{1:n}$.
One could select these arbitrarily or by grid search but this would be wasteful;
uniformly random search has also been proposed as a surprisingly good alternative
to grid search~\cite{Bergstra:2012}.
There is, however, a rich literature on selection strategies that utilize the
surrogate model to guide the sequential search, \emph{i.e.} the selection of the
next query point $\vx_{n+1}$ given $\D_n$.

The key idea behind these strategies is to define an acquisition functions
$\alpha:\R^d\mapsto\R$ which quantifies the promise\footnote{The way ``promise''
is quantified depends on whether we care about cumulative losses or only the
final recommendation $\xrec_N$.} of any point in the search space.
The acquisition function is carefully designed to trade off exploration of the
search space and exploitation of current promising neighborhoods.
There are three common types of acquisition functions: improvement-based,
optimistic, and information-based policies.

The improvement-based acquisition functions, probability and expected improvement
(PI and EI, respectively), select the next point
with the most probable and most expected improvement,
respectively~\cite{Kushner:1964,Mockus:1978}.
On the other hand, the optimistic policy \emph{upper confidence bound} (GP-UCB)
measures, marginally for each test point $\vx$, how good the corresponding
observation $y$ will be in a low and fixed probability ``good case scenario''---hence 
the optimism~\cite{Srinivas:2010}.
In contrast, there exist information-based methods such as randomized
probability matching, also known as
Thompson sampling~\cite{Thompson:1933,Scott:2010,Li:2011,Kaufmann:2012,Agrawal:2013},
or the more recent \emph{entropy search}
methods~\cite{Villemonteix:2009,Hennig:2012,Hernandez:2014}.
Thompson sampling selects the next point according to the distribution of the
optimum $\xstar$, which is induced by the current
posterior~\cite{Scott:2010,Hernandez:2014,Shahriari:2014}.
Meanwhile, entropy search methods select the point $\vx$ that is expected to
provide the most information towards reducing uncertainty about $\xstar$.

In this work we focus our attention on EI, which is perhaps the most common
acquisition function. With the GP surrogate model, EI has the following analytic
expression
\begin{equation}
	\alpha^{\textsc{EI}}_n(\vx)
	= (\mu_n(\vx) - \tau) \Phi\left(\frac{\mu_n(\vx) - \tau}{\sigma_n(\vx)}\right)
	+ \sigma_n(\vx)\phi\left(\frac{\mu_n(\vx) - \tau}{\sigma_n(\vx)}\right),
	\label{equ:ei}
\end{equation}
where $\Phi$ and $\phi$ denote the standard normal cumulative distribution
and density functions, respectively.
Note however that the technique we outline in the next section
can readily be extended to any Gaussian process derived acquisition function,
including all those mentioned above.

\section{Unbounded Bayesian optimization}
\label{sec:unbounded}

\subsection{Volume doubling}
\label{sec:doubling}

Our first proposed heuristic approach consists in expanding the search space
regularly as the optimization progresses, starting with an initial user-defined
bounding box.
This method otherwise follows the standard Bayesian optimization procedure and
optimizes within the bounding box that is available at the given time step $n$.
This approach requires two parameters: the number of iterations between
expansions and the growth factor $\gamma$.
Naturally, to avoid growing the feasible space $\calX$ by a factor that is
exponential in $d$, the growth factor applies to the volume of $\calX$.
Finally, the expansion is isotropic about the centre of the domain.
In this work, we double ($\gamma=2$) the volume every $3d$ evaluations
(only \emph{after} an initial latin hypercube sampling of $3d$ points).

\subsection{Regularizing improvement policies}
\label{sec:regularizing}

We motivate the regularized approach by considering improvement policies
(\emph{e.g.}\ EI); however, in the next section we show that this proposed
approach can be applied more generally to all GP-derived acquisition functions.
Improvement policies are a popular class of acquisition functions that rely on
the improvement function
\begin{equation}
	I(\x) = (f(\x) - \tau)\I[f(\x) \geq \tau]
	\label{eq:improvement}
\end{equation}
where $\tau$ is some target value to improve upon.
Expected improvement then compute $\E[I(\x)]$ under the posterior GP.

When the optimal objective value $\fstar$ is known and we set $\tau=\fstar$,
these algorithms are referred to as \emph{goal seeking}~\cite{Jones:2001}.
When the optimum is not known, it is common to use a proxy for $\fstar$ instead,
such as the value of the best observation so far, $y^+$; or in the noisy setting, one
can use either the maximum of the posterior mean or the value of the posterior
at the \emph{incumbent} $\x^+$, where $\x^+\in\argmax_{\x\in\x_{1:n}} \mu_n(\x)$.

In some cases, the above choice of target $\tau = y^+$ can lead to a lack of
exploration, therefore it is common to choose a \emph{minimum improvement}
parameter $\xi>0$ such that $\tau =(1 + \xi)y^+$ (for convenience here we assume
$y^+$ is positive and in practice one can subtract the overall mean to make it so).
Intuitively, the parameter $\xi$ allows us to require a minimum fractional
improvement over the current best observation $y^+$.
Previously, the parameter $\xi$ had always been chosen to be constant if not zero.
In this work we propose to use a function $\xi: \R^d\mapsto\R^+$ which maps
points in the space to a value of fractional minimum improvement.
Following the same intuition, the function $\xi$ lets us require larger expected
improvements from points that are \emph{farther} and hence acts as a regularizer
that penalizes distant points. The improvement function hence becomes:
\begin{equation}
  I(\x) = (f(\x) - \tau(\x))\I[f(\x) \geq \tau(\x)],\quad\text{with}\quad
  \tau(\x) = (1 + \xi(\x))y^+,
  \label{equ:nonstat-imp}
\end{equation}
where the choice of $\xi(\x)$ is discussed in Section~\ref{sec:choice-reg}.

\subsubsection{Extension to general policies}
\label{sec:prior-mean}

In the formulation of the previous section, our method seems restricted to
improvement policies. However, many recent acquisition functions of interest are
not improvement-based, such as GP-UCB, entropy search, and Thompson sampling.
In this section, we describe a closely related formulation that
generalizes to all acquisition functions that are derived from a GP surrogate
model.

\begin{figure}
    \centering
	\begin{subfigure}[b]{0.45\textwidth}
	    \includegraphics[width=\textwidth]{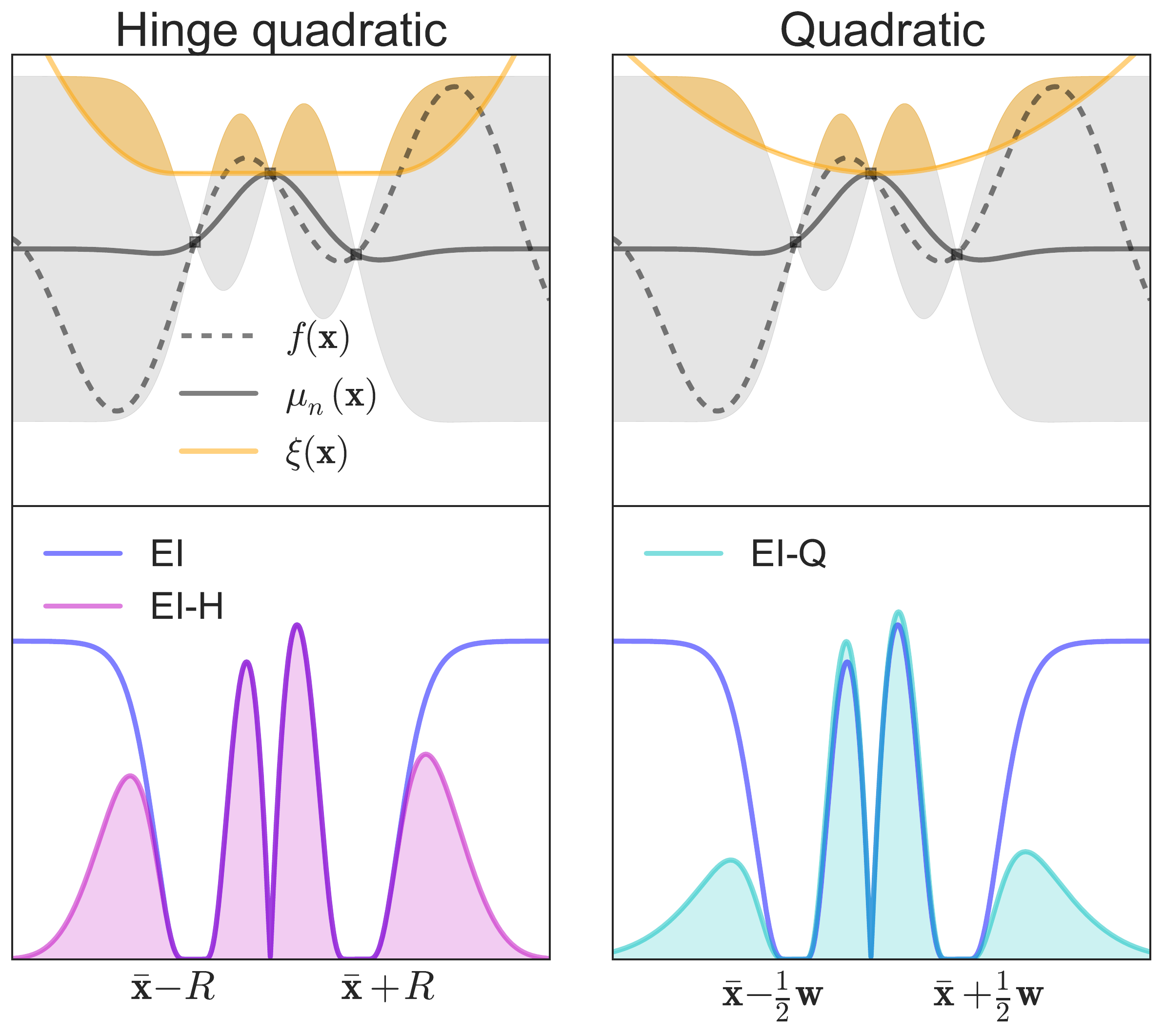}
        \caption{Minimum improvement view}
		\label{fig:minimp}
    \end{subfigure}
	\hspace{2em}
	\begin{subfigure}[b]{0.45\textwidth}
    	\includegraphics[width=\textwidth]{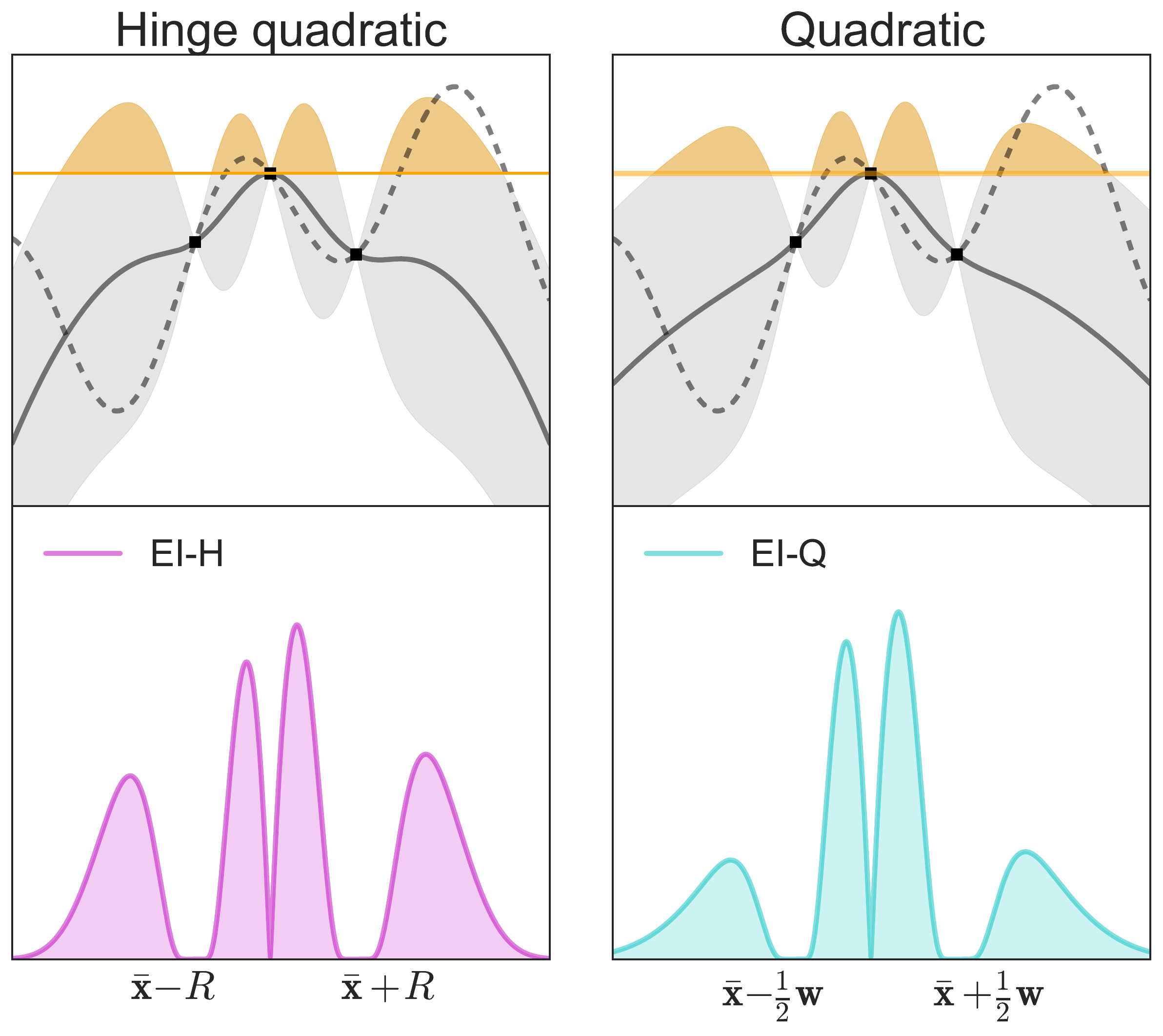}
        \caption{Prior mean view}
		\label{fig:priormean}
    \end{subfigure}
    \caption{\small
		Visualization of the two alternate views of regularization in
		Bayesian optimization.
		The objective function and posterior surrogate mean are represented
		as black dashed and solid lines, respectively, with grey shaded areas
		indicating $\pm2\sigma_n$.
		Integrating the surrogate model above the target (orange shaded area)
		results in the regularized EI acquisition function (magenta and cyan).
		Using a non-stationary target with a constant prior mean (left) or
		a fixed target with a non-stationary prior mean (right) lead to
		indistinguishable acquisition functions, which decay at infinity.}
    \label{fig:viz-minimp}
\end{figure}

Consider expanding our choice of non-stationary target $\tau$ in
Equation~\eqref{equ:nonstat-imp}
\begin{align}
	I(\x) &= (f(\x) - y^+(1 + \xi(\x)))\I[f(\x) \geq y^+(1 + \xi(\x))]
	\notag
	\\
	&= (f(\x) - y^+\xi(\x) - y^+)\I[f(\x) - y^+\xi(\x) \geq y^+]
	\notag
	\\
	&= (\tilde f(\x) - y^+) \I[\tilde f(\x) \geq y^+]
	\label{eq:reg-improvement}
\end{align}
where $\tilde f$ is the posterior mean of a GP from \eqref{eq:gpmean} with
prior mean $\tilde\mu_0(\x) = \mu_0(\x) - y^+\xi(\x)$. Notice the similarity
between \eqref{eq:improvement} and \eqref{eq:reg-improvement}. Indeed, in its
current form we see that the regularization can be achieved simply by using a
different prior mean $\tilde\mu_0$ and a constant target $y^+$. This duality can
be visualized when comparing the left and right panel of
Figure~\ref{fig:viz-minimp}.

Precisely speaking, these two views are not exactly equivalent. Starting with
the $\tilde\mu_0$ prior mean, the posterior mean yields an additional term
\begin{equation}
	\vk(\vx)\T(\vK+\sigma^2\vI)^{-1}\xi(\vX),
\end{equation}
where $[\xi(\vX)]_i = \xi(\vx_i)$.
This term is negligible when the test point $\vx$ is far from the data $\vX$
due to our exponentially vanishing kernel, making the two alternate views
virtually indistinguishable in Figure~\ref{fig:viz-minimp}.
However, with this new formulation we can apply the
same regularization to any policy which uses a Gaussian process,
\emph{e.g.}\ Thompson sampling or entropy search.

\subsubsection{Choice of regularization}
\label{sec:choice-reg}

By inspecting Equation~\eqref{equ:ei}, it is easy to see that any \emph{coercive}
prior mean function would lead to an asymptotically vanishing EI acquisition
function; in this work however we consider quadratic (Q) and
isotropic hinge-quadratic (H) regularizing prior mean functions, defined as
follows (excluding the constant bias $b$)
\begin{align}
	\xi_Q(\x) &= (\x - \bar\x)\T \diag(\vw^2)^{-1} (\x - \bar\x),
    \label{eq:reg-quad}
    \\
    \xi_H(\x) &= \mathbb{I} [\Vert\x - \bar\x\Vert_2 > R] \frac{\Vert\x - \bar\x\Vert_2 - R}
                                                        {\beta R}.
    \label{eq:reg-hinge}
\end{align}
Both of these regularizers are parameterized by $d$ location parameters
$\bar\vx$, and while $\xi_Q$ has an additional $d$ width parameters $\vw$,
the isotropic regularizer $\xi_H$ has a single radius $R$ and a single
$\beta$ parameter, which controls the curvature of $\xi_H$ outside the ball
of radius $R$; in what follows we fix $\beta=1$.

\paragraph{Fixed prior mean hyperparameters.}
We are left with the choice of centre $\bar\vx$ and radius parameters $R$ (or
widths $\vw$).
Unlike the bias term $b$, these parameters of the regularizer are not intended
to allow the surrogate to better fit the observations $\D_n$.
In fact, using the marginal likelihood to estimate or marginalize
$\psi=\{\bar\vx, R, \vw\}$, could lead to fitting the regularizer to a local
mode which could trap the algorithm in a suboptimal well. For this reason, we
use an initial, temporary user-defined bounding box to set $\psi$ at the
beginning of the run; the value of $\psi$ remains fixed in all subsequent
iterations.

%We choose a data dependent regularizer in order to let our
%observations guide the future areas of interest. As is already done with other
%model hyper-parameters (\emph{e.g.,} kernel length scales and variance) we
%prescribe a log-normal prior on $R$ such that
%\begin{align}
%    \bar \x &\sim \Normal(\mu_{\bar \x}, \Sigma_{\bar \x}),
%    \\
%    R &\sim \ln\Normal(\mu_R, \sigma^2_R),
%    \label{eq:w-prior}
%\end{align}
%and use Markov chain Monte Carlo (MCMC) to sample from the unnormalized
%posterior
%\begin{equation}
%    p(\bar \x, R | \D_t) \propto p(\D_t | \bar \x, R) p(\bar \x) p(R), 
%\end{equation}
%where $p(\D_t | \vw)$ is the GP marginal likelihood which can be computed
%analytically.
%
%When sampling the parameters of the quadratic prior mean $\hat\mu_Q$ via Monte
%Carlo techniques mentioned above, we run the risk of fitting a local optimum of
%the response surface. The quadratic prior mean is especially prone to this
%overfitting and could stifle exploration. Meanwhile, the hinge
%quadratic $\hat\mu_H$ has the desired feature that within a radius $R$ from the
%initial window, the regularizer has no effect and simply fits a constant bias
%$\mu_0$ which corresponds to current Bayesian optimization standard practice.

\subsection{Visualization}

\begin{figure}
	\centering
	\includegraphics[width=0.45\textwidth]{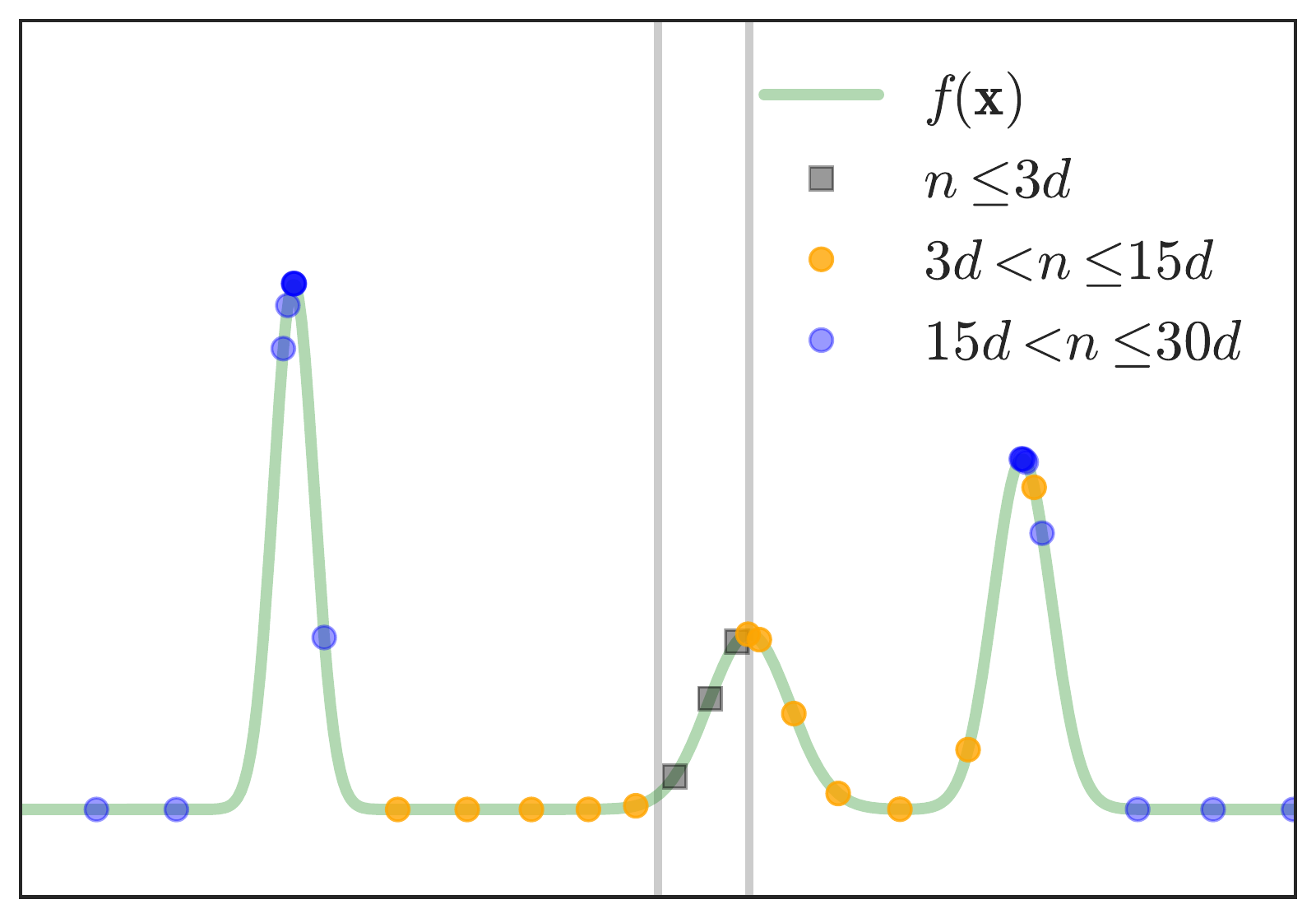}
	\includegraphics[width=0.31\textwidth]{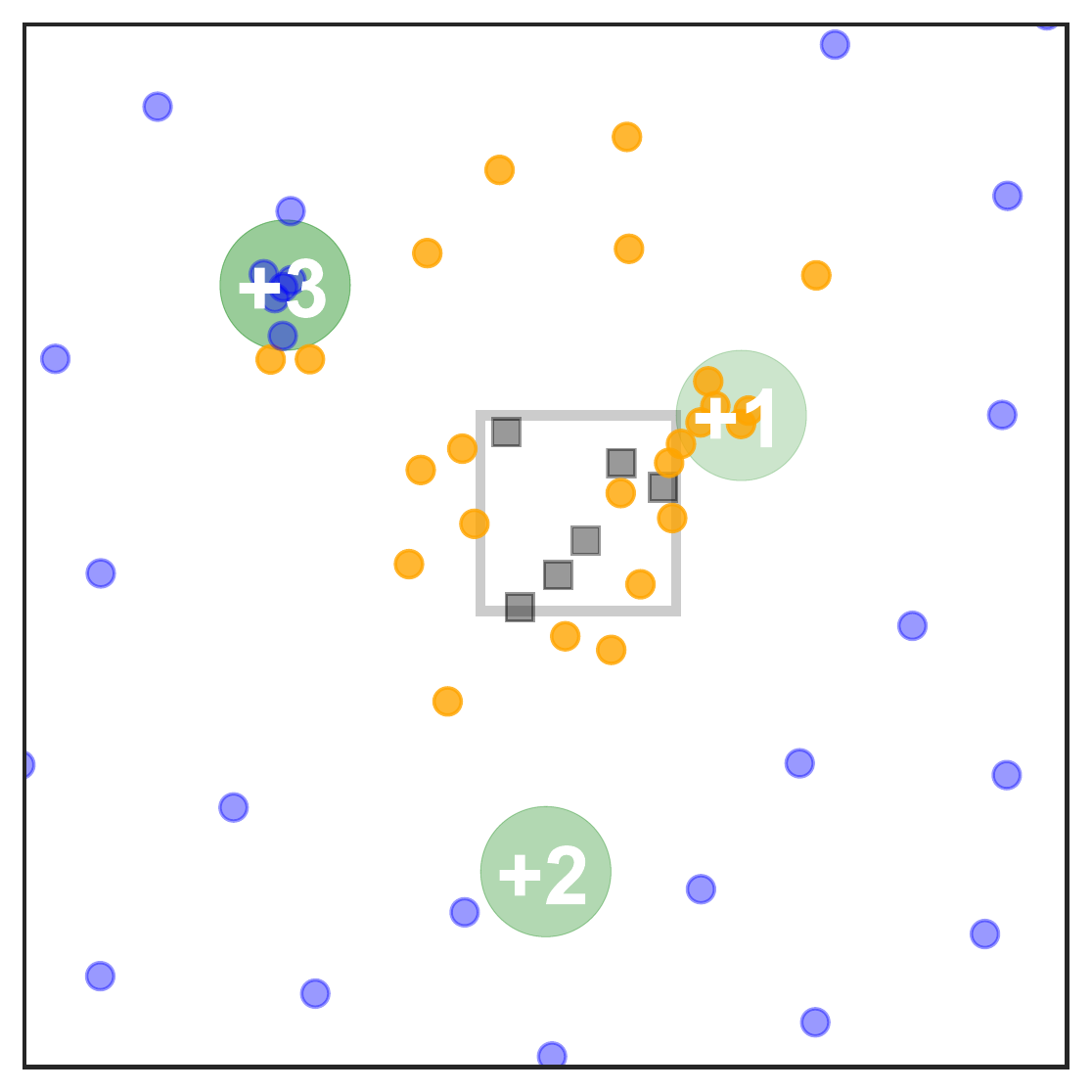}
	\caption{\small
		Visualization of selections $\vx_n$ made by the EI-H algorithm on two
		toy objective functions consisting of
		three Gaussian modes in one (left) and two (right) dimensions.
		Grey lines delimit the initial bounding box;
		grey square markers indicate the initial latin hypercube points,
		while the orange and blue points distinguish between the first and
		second half of the evaluation budget of $30d$, respectively.
		In the two-dimensional example, the height of the Gaussians are
		indicated by +1, +2, and +3.}	
  \label{fig:visualization}
\end{figure}

Before describing our experimental results, Figure~\ref{fig:visualization}
provides a visualization of EI with the hinge-quadratic prior mean, optimizing
two toy problems in one and two dimensions.
The objective functions consist of three Gaussian modes of varying heights and
the initial bounding box does not include the optimum.
We draw attention to the way the space is gradually explored outward from the
initial bounding box.

%In contrast, between $t=10$ and $t=15$, the right-most Gaussian mode was queried 4 times and was able
%to resolve the peak. By $t=18$, the uncertainty about the right-most peak is so
%low that EI-H vanishes and new green modes appear to left and right side. These
%new lobes lead to further exploration and finally the optimal peak is found.
%Once again, at $t=20$ the newly found peak is resolved via multiple querying.

%===============================================================================
\section{Experiments}
\label{sec:experiments}

In this section, we evaluate our proposed methods and show that they achieve
the desirable behaviour on two synthetic benchmarking functions, and  a simple task of tuning the stochastic gradient descent and regularization parameters
used in training a multi-layered perceptron (MLP) and a convolutional neural
network (CNN) on the MNIST dataset.

\paragraph{Experimental protocol.}
For every test problem of dimension $d$ and every algorithm, 
the optimization was run with an overall evaluation budget of $30d$ including an
initial $3d$ points sampled according to a latin hypercube sampling scheme (as
suggested in~\cite{Jones:2001}).
Throughout each particular run, at every iteration $n$ we record the value of
the best observation up to $n$ and report these in Figure~\ref{fig:results}.

\paragraph{Algorithms.}
We compared the two different methods from Section~\ref{sec:unbounded} to the
standard EI with a fixed bounding box.
Common random seeds were used for all methods in order to reduce confounding
noise.
All algorithms were implemented in the \texttt{pybo} framework available on
github\footnote{
https://github.com/mwhoffman/pybo
}~\cite{Hoffman:2015}, and are labelled in our plots as follows:

\begin{description}
\item[EI:]
	Vanilla expected improvement with hyperparameter marginalization.
\item[EI-V:]
	Expected improvement with the search volume doubled every $3d$ iterations.
\item[EI-H/Q:]
	Regularized EI with a hinge-quadratic and quadratic prior mean, respectively.
%\item[RS:]
%	Random selections uniformly sampled within the user-defined bounding box.
\end{description}

Note that for the regularized methods EI-H/Q, the initial bounding box is only
used to fix the location and scale of the regularizers, and to sample initial
query points. In particular, both regularizers are centred around the box centre;
for the quadratic regularizer the width of the box in each direction is used
to fix $\vw$, whereas for the hinge-quadratic $R$ is set to the box circumradius.
Once these parameters are fixed, the bounding box is no longer relevant.

\subsection{Synthetic benchmarks: Hartmann 3 and 6}

The Hartmann 3 and 6 functions (numbers refer to their dimensionality) are
standard, synthetic global optimization benchmarking test functions.
In a separate experiment, indicated by an appended asterisk ($\ast$), we consider
a randomly located bounding box of side length 0.2 within the unit hypercube.
This window is unlikely to include the global optimum, especially in
the six-dimensional problem, which makes this a good problem to test whether
our proposed methods are capable of useful exploration outside the initial
bounding box. 

\subsection{MLP and CNN on MNIST}

The MNIST hand-written digit recognition dataset is a very common simple task for
testing neural network methods and architectures. In this work we optimize the
parameters of the stochastic gradient descent (SGD) algorithm used in training
the weights of the neural network. In particular, we consider an MLP with 2048
hidden units with tanh non-linearities, and a CNN with two convolutional layers.
These examples were taken from the official GitHub repository of \texttt{torch7}
demos\footnote{https://github.com/torch/demos}. The code written for this work
can be readily extended to any other demo in the repository or any torch script.

For this problem, we optimize four parameters, namely the learning rate and
momentum of the SGD optimizer, and the $L_1$ and $L_2$ regularization
coefficients.
The parameters were optimized in log space (base $e$) with an initial bounding
box of $[-3, -1] \times [-3, -1] \times [-3, 1] \times [-3, 1]$, respectively.
For each parameter setting, a black-box function evaluation corresponds to
training the network for one epoch and returning the test set accuracy.
To be clear, the goal of this experiment is not to achieve state-of-the-art for
this classification task but instead to demonstrate that our proposed
algorithms can find optima well outside their initial bounding boxes.

\subsection{Results}

\begin{figure}
	\centering
	\includegraphics[width=0.99\textwidth]{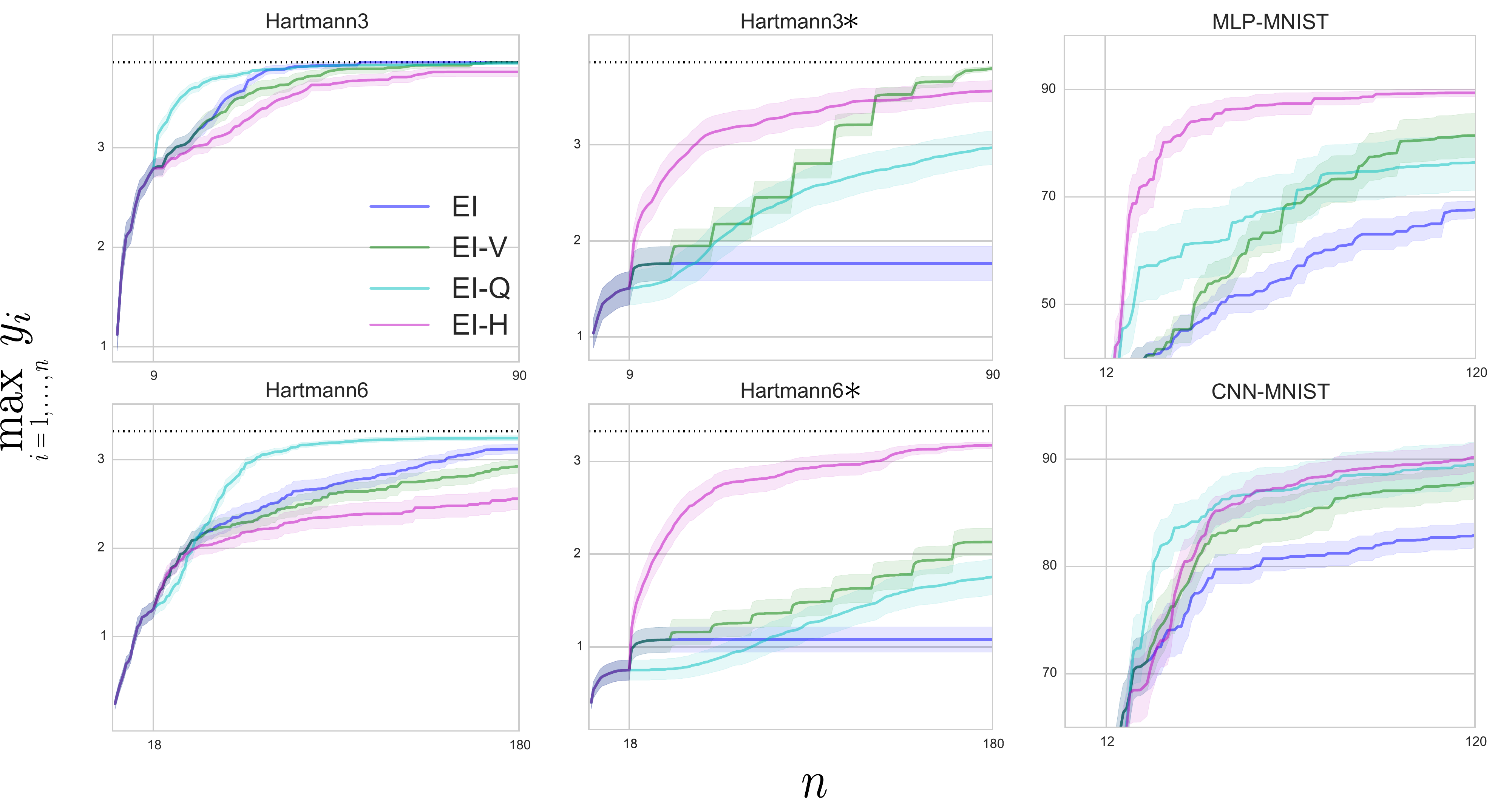}
	\caption{\small
		Best observations as optimization progresses. Plotting mean and
		standard error over 40 (Hartmann), 14 (MLP), and 19 (CNN) repetitions.}
	\label{fig:results}
\end{figure}

Figure~\ref{fig:results} shows that for the Hartmann tests, the Bayesian optimization approaches proposed in this paper work well. The results confirm our hypothesis that the proposed methods are capable of useful exploration outside the initial bounding box. We note that when using the entire unit hypercube as the initial box, all the Bayesian optimization techniques exhibit similar performance as in this case the optimum is within the box.
The Hartmann tests also show that our volume doubling heuristic is a good baseline method. Although it is less effective than EI-H as the dimensionality increases, it is nonetheless an improvement over standard EI in all cases.

The MNIST experiment shows good performance from all three methods EI-\{V,H,Q\},
particularly from the hinge-quadratic regularized algorithm. Indeed, when compared
to the standard EI, EI-H boasts a 20\% improvement in accuracy on the MLP and almost
10\% on the CNN.

\begin{figure}
	\centering
	\includegraphics[width=0.6\textwidth]{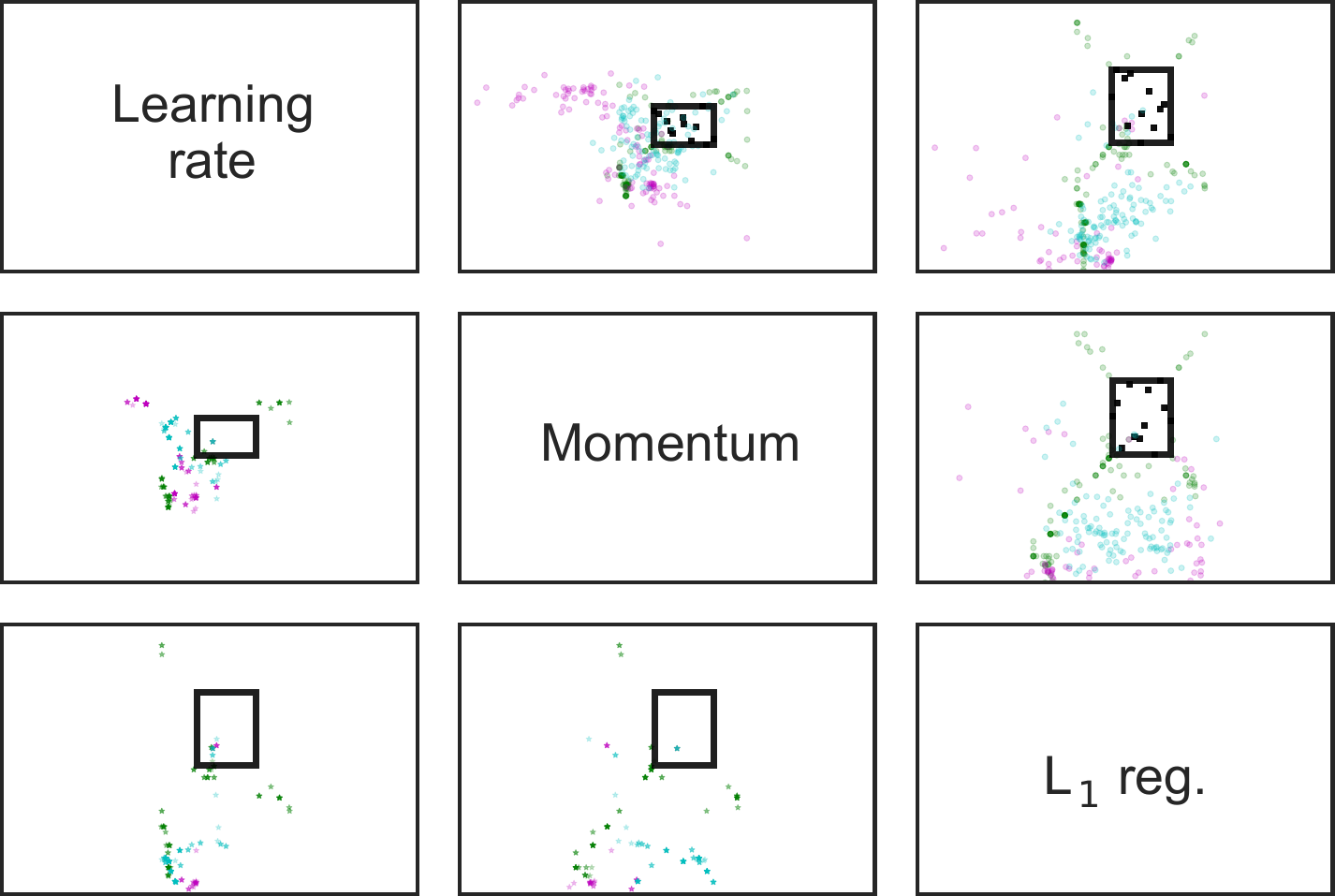}
	\caption{\small
		Pairwise scatterplot of selections (upper triangle) and
		recommendations (lower triangle) for the MLP-MNIST experiment.
		For example, the second plot of the first row corresponds to a scatter
		plot of the selected learning rates vs. momentum parameters \emph{for a
		single seed}.
		In contrast, the first plot of the second row corresponds to a scatter
		plot of the recommended learning rates and momentum parameters
		\emph{over all 14 runs}.
		The initial bounding box and sample points (for this particular run) are
		shown as a black rectangle and black square dots, respectively.
		All other points respect the color scheme of Figure~\ref{fig:results}.
		($L_2$ cropped out for space.)}
	\label{fig:selections}
\end{figure}

We believe that EI-H performs particularly well in settings where a small initial
bounding box is prescribed because the hinge-quadratic regularizer allows the
algorithm to explore outward more quickly. In contrast, EI-Q performs better when
the optimum is included in the initial box; we suspect that this is due to the fact
that the regularizer avoids selecting boundary and corner points, which EI and EI-V
tend to do, as can be seen in Figure~\ref{fig:selections}.
Indeed, the green dots (EI-V) follow the corners of the growing bounding box.
In contrast, EI-H/Q do not exhibit this artefact.

%===============================================================================
\section{Conclusion and future work}
\label{sec:conclusion}

In this work, we propose a versatile new approach to Bayesian optimization which
is not limited to a search within a bounding box.
Indeed, given an initial bounding box that does not include the optimum, we have
demonstrated that our approach can expand its region of interest and achieve
greater function values.
Our method fits seamlessly within the current Bayesian optimization framework,
and can be readily used with any acquisition function which is induced by a GP.  

Finally, this could have implications in a distributed Bayesian optimization
scheme whereby multiple Bayesian optimization processes are launched, each
responsible for a neighbourhood of the domain.
Interactions between these processes could minimize overlaps or alternatively
ensure a certain level of redundancy, which would be helpful in a noisy setting.

We emphasize that in this work we have addressed one of the challenges that must be overcome toward the development of practical Bayesian optimization hyper-parameter tuning tools. A full solution, however, must also address the issues of dimensionality, non-stationarity, and early stopping. Fortunately, there has been great recent progess along these directions, and the methods proposed in this paper provide a complementary and essential piece of the puzzle.

%===============================================================================
\bibliographystyle{plain}
%\bibliography{bayesopt}
\bibliography{rei}

\begin{thebibliography}{10}

\bibitem{Agrawal:2013}
S.~Agrawal and N.~Goyal.
\newblock Thompson sampling for contextual bandits with linear payoffs.
\newblock In {\em International Conference on Machine Learning}, 2013.

\bibitem{Bergstra:2011}
J.~Bergstra, R.~Bardenet, Y.~Bengio, and B.~K{\'e}gl.
\newblock Algorithms for hyper-parameter optimization.
\newblock In {\em Advances in Neural Information Processing Systems}, pages
  2546--2554, 2011.

\bibitem{Bergstra:2012}
J.~Bergstra and Y.~Bengio.
\newblock Random search for hyper-parameter optimization.
\newblock {\em Journal of Machine Learning Research}, 13:281--305, 2012.

\bibitem{Li:2011}
O.~Chapelle and L.~Li.
\newblock An empirical evaluation of {Thompson} sampling.
\newblock In {\em Advances in Neural Information Processing Systems}, pages
  2249--2257, 2011.

\bibitem{Hennig:2012}
P.~Hennig and C.~J. Schuler.
\newblock Entropy search for information-efficient global optimization.
\newblock {\em The Journal of Machine Learning Research}, pages 1809--1837,
  2012.

\bibitem{Hernandez:2014}
J.~M. Hern\'andez-Lobato, M.~W. Hoffman, and Z.~Ghahramani.
\newblock Predictive entropy search for efficient global optimization of
  black-box functions.
\newblock In {\em Advances in Neural Information Processing Systems}. 2014.

\bibitem{Hoffman:2015}
M.~W. Hoffman and B.~Shahriari.
\newblock Modular mechanisms for {B}ayesian optimization.
\newblock In {\em NIPS workshop on {B}ayesian Optimization}, 2014.

\bibitem{Hoffman:2014}
M.~W. Hoffman, B.~Shahriari, and N.~de~Freitas.
\newblock On correlation and budget constraints in model-based bandit
  optimization with application to automatic machine learning.
\newblock In {\em AI and Statistics}, pages 365--374, 2014.

\bibitem{Hutter:2010b}
F.~Hutter, T.~Bartz-Beielstein, H.~H. Hoos, K.~Leyton-Brown, and K.~P. Murphy.
\newblock Sequential model-based parameter optimisation: an experimental
  investigation of automated and interactive approaches.
\newblock In T.~Bartz-Beielstein, M.~Chiarandini, L.~Paquete, and M.~Preuss,
  editors, {\em Empirical Methods for the Analysis of Optimization Algorithms},
  chapter~15, pages 361--411. Springer, 2010.

\bibitem{Jones:2001}
D.~R. Jones.
\newblock A taxonomy of global optimization methods based on response surfaces.
\newblock {\em J. of Global Optimization}, 21(4):345--383, 2001.

\bibitem{Kaufmann:2012}
E.~Kaufmann, N.~Korda, and R.~Munos.
\newblock Thompson sampling: An asymptotically optimal finite-time analysis.
\newblock In {\em Algorithmic Learning Theory}, volume 7568 of {\em Lecture
  Notes in Computer Science}, pages 199--213. Springer Berlin Heidelberg, 2012.

\bibitem{Kushner:1964}
Harold~J Kushner.
\newblock A new method of locating the maximum point of an arbitrary multipeak
  curve in the presence of noise.
\newblock {\em Journal of Fluids Engineering}, 86(1):97--106, 1964.

\bibitem{Mahendran:2012}
N.~Mahendran, Z.~Wang, F.~Hamze, and N.~{de Freitas}.
\newblock Adaptive {MCMC} with {Bayesian} optimization.
\newblock {\em Journal of Machine Learning Research - Proceedings Track},
  22:751--760, 2012.

\bibitem{Mockus:1978}
Jonas Mo{\v c}kus, V~Tiesis, and Antanas {\v Z}ilinskas.
\newblock The application of bayesian methods for seeking the extremum.
\newblock In L~Dixon and G~Szego, editors, {\em Toward Global Optimization},
  volume~2. Elsevier, 1978.

\bibitem{Rasmussen:2006}
C.~E. Rasmussen and C.~K.~I. Williams.
\newblock {\em Gaussian Processes for Machine Learning}.
\newblock The MIT Press, 2006.

\bibitem{Scott:2010}
S.~L. Scott.
\newblock A modern {B}ayesian look at the multi-armed bandit.
\newblock {\em Applied Stochastic Models in Business and Industry},
  26(6):639--658, 2010.

\bibitem{Shahriari:2014}
B.~Shahriari, Z.~Wang, M.~W. Hoffman, A.~Bouchard-C\^ot\'e, and N.~de~Freitas.
\newblock An entropy search portfolio.
\newblock In {\em NIPS workshop on {Bayesian} Optimization}, 2014.

\bibitem{Snoek:2012}
J.~Snoek, H.~Larochelle, and R.~P. Adams.
\newblock Practical {Bayesian} optimization of machine learning algorithms.
\newblock In {\em Advances in Neural Information Processing Systems}, pages
  2951--2959, 2012.

\bibitem{Snoek:2015DNGO}
J.~Snoek, O.~Rippel, K.~Swersky, R.~Kiros, N.~Satish, N.~Sundaram, M.~Patwary,
  M.~Ali, R.~P. Adams, et~al.
\newblock Scalable {B}ayesian optimization using deep neural networks.
\newblock {\em arXiv preprint arXiv:1502.05700}, 2015.

\bibitem{Srinivas:2010}
N.~Srinivas, A.~Krause, S.~M. Kakade, and M.~Seeger.
\newblock Gaussian process optimization in the bandit setting: No regret and
  experimental design.
\newblock In {\em International Conference on Machine Learning}, pages
  1015--1022, 2010.

\bibitem{Swersky:2013}
K.~Swersky, J.~Snoek, and R.~P. Adams.
\newblock Multi-task {Bayesian} optimization.
\newblock In {\em Advances in Neural Information Processing Systems}, pages
  2004--2012, 2013.

\bibitem{Thompson:1933}
W.~R. Thompson.
\newblock On the likelihood that one unknown probability exceeds another in
  view of the evidence of two samples.
\newblock {\em Biometrika}, 25(3/4):285--294, 1933.

\bibitem{Villemonteix:2009}
J.~Villemonteix, E.~Vazquez, and E.~Walter.
\newblock An informational approach to the global optimization of
  expensive-to-evaluate functions.
\newblock {\em J. of Global Optimization}, 44(4):509--534, 2009.

\end{thebibliography}

%===============================================================================
\end{document}